\documentclass[10pt,letterpaper]{article}
\usepackage[utf8]{inputenc}
\usepackage[T1]{fontenc}
\usepackage[english]{babel}
\usepackage{microtype}
\usepackage{graphicx}
\usepackage{booktabs,multirow,array,tabularx}
\usepackage{amsmath,amssymb}
\usepackage{float}
\usepackage{capt-of}
\usepackage{enumitem}
\usepackage{xcolor}
\usepackage{tikz}
\usetikzlibrary{arrows.meta,positioning,fit}
\usepackage{url}
\usepackage{listings}
\usepackage[numbers,sort&compress]{arxiv/natbib}
\usepackage{hyperref}
\usepackage{arxiv/ssArxiv}

\definecolor{wacvblue}{rgb}{0.21,0.49,0.74}
\newcommand{\method}{LexiDet-OVD}
\newcommand{\coco}{COCO1000}
\newcommand{\lvis}{LVIS1000-30c}
\newcommand{\pc}{Prompt Consensus}

\newcommand{\apfifty}{AP$_{50}$}
\newcommand{\best}[1]{\textbf{#1}}

\title{LexiDet-OVD: Benchmarking Lexical Sensitivity in Open-Vocabulary Object Detection}
\author{Amit Thakur \qquad Mukesh Singhal \\
University of California, Merced, USA \\
\texttt{athakur5@ucmerced.edu} \qquad \texttt{msinghal@ucmerced.edu} \\
}
\hypersetup{pdftitle={LexiDet-OVD: Benchmarking Lexical Sensitivity in Open-Vocabulary Object Detection},
  pdfauthor={Amit Thakur and Mukesh Singhal},
  pdfsubject={Accepted WACV 2027 Datasets Track paper; arXiv preprint}}

\begin{document}
\maketitle
\vspace{-25pt}
\begingroup\renewcommand{\thefootnote}{}\footnotetext{\textcopyright\ 2027 IEEE. Personal use of this material is permitted. Permission from IEEE must be obtained for all other uses, in any current or future media, including reprinting/republishing this material for advertising or promotional purposes, creating new collective works, for resale or redistribution to servers or lists, or reuse of any copyrighted component of this work in other works.}\endgroup

\begin{abstract}
Open-vocabulary object detectors localize objects from free-form text prompts,
but their reliability depends on how users describe the target. Standard
evaluations usually test a fixed class name, leaving prompt-induced failure
modes underexplored. We introduce \method, a benchmark and metric suite for
studying lexical sensitivity in open-vocabulary object detection. We organize
the prompt sets into equivalent, conditional, and broad semantic scopes, and
evaluate four detector families on fixed COCO and LVIS samples, each covering 30 classes
(800 held-out COCO images and 1,000 LVIS images). Beyond mean and
worst-prompt \apfifty, we report P10, CVaR$_{20}$, leave-one-prompt-out
influence, detection agreement, box consistency, and confidence variance using
one-to-one multi-instance matching. Equivalent-prompt best-to-worst gaps range
from 5.3 to 23.4 points on \coco\ and from 5.1 to 29.9 points on \lvis, showing
that strong average accuracy and lexical stability are distinct properties. We
compare maximum-score selection, weighted-box fusion, \pc, and an exploratory
fusion rule. Aggregation recovers performance in several settings but no method
dominates all model--dataset pairs. A controlled Detic/SHiNe comparison further
shows that semantic hierarchy enrichment does not by itself eliminate lexical
sensitivity. These results support reporting prompt-distribution robustness
alongside conventional open-vocabulary detection accuracy.
\end{abstract}

\section{Introduction}
Open-vocabulary object detection (OVD) has made object localization
increasingly interactive: a user can type a phrase such as ``bicycle'',
``person'', or ``traffic light'', and a detector returns bounding boxes for the
requested concept. This ability is enabled by vision--language pre-training and
language-conditioned architectures such as OWL-ViT~\cite{minderer2022owlvit},
GLIP~\cite{li2022glip}, Detic~\cite{zhou2022detic}, YOLO-World~\cite{cheng2024yoloworld},
RegionCLIP~\cite{zhong2022regionclip}, ViLD~\cite{gu2022vild}, and Grounding
DINO~\cite{liu2023groundingdino}. Because the user interface is language,
deployment reliability depends not only on visual recognition but also on the
exact wording used at inference time.

Most OVD evaluations ask whether a model can detect a fixed category name. Yet
real users rarely know the canonical vocabulary expected by a benchmark or
detector. One user may ask for ``bicycle'', another for ``bike'', another for
``bicycles'', and another for ``a photo of a bicycle''. A reliable
language-conditioned detector should not fail under ordinary rewordings that
preserve the intended visual concept. At the same time, ``vehicle'' and
``bicycle on the road'' are not interchangeable with ``bicycle'': the first is
broader and the second is conditional on the image. A useful benchmark must
therefore vary language without silently changing the evaluation target.

This paper studies \emph{lexical sensitivity in OVD}: the degree to which
detections change when a fixed target class is described using different
natural-language prompts. Prompt design has been studied for vision--language
models~\cite{radford2021clip,jia2021align,zhou2022coop,zhou2022cocoop}, but a
detection benchmark must measure more than a class score. A wording change can
remove a detection, move its box to another instance, preserve localization but
change confidence, or produce boxes for an absent category. These failure modes
are hidden when evaluation reports only one canonical prompt.

We introduce \method, a controlled benchmark and metric suite for auditing this
behavior. It holds the image, intended class, annotations, model, and threshold
fixed while varying an audited prompt realization. The primary protocol uses
only semantically equivalent prompts. Conditional expressions and broad
taxonomic terms remain valuable stress tests, but are reported separately so a
correct response to a broader query is not mislabeled as lexical instability.
We evaluate Grounding DINO, OWL-ViT, OWLv2, and YOLO-World-v2 on \coco\ and
\lvis. This two-dataset, four-family design tests whether the phenomenon
persists across architectures, model generations, and visual vocabularies.

\noindent\textbf{Contributions:} We make four contributions:
\begin{itemize}[leftmargin=*,nosep]
  \item a deterministic two-dataset benchmark with versioned prompt manifests,
  programmatic audit decisions, and explicit semantic-scope labels;
  \item a metric suite combining mean, tail, leave-one-out, detection-state,
  localization, and confidence statistics with multi-instance matching;
  \item an evaluation across four detector families and two datasets showing that lexical
  sensitivity persists even among equivalent prompts and varies by model and dataset;
  and
  \item matched comparisons of four training-free aggregation rules and a
  controlled Detic/SHiNe study, with statistical and computational diagnostics.
\end{itemize}

Our findings do not support a universal mitigation claim. Prompt aggregation
and hierarchy-aware classifiers help selected settings, but neither removes
sensitivity consistently. The central conclusion is methodological:
prompt selection is part of the evaluated system, and OVD results should expose
the performance distribution induced by plausible equivalent wording.

\section{Related Work}
\textbf{Open-vocabulary object detection:}
OVD extends detection beyond a fixed training vocabulary by connecting visual
localization to text embeddings, phrase grounding, or class-agnostic proposal
mechanisms. ViLD and RegionCLIP transfer vision--language knowledge to region
recognition~\cite{gu2022vild,zhong2022regionclip}; Detic scales detector
vocabulary using image-level labels~\cite{zhou2022detic}; and GLIP and
Grounding DINO formulate detection as grounded language understanding
~\cite{li2022glip,liu2023groundingdino}. OWL-ViT transfers contrastive
image--text representations to detection, OWLv2 scales that design, and
YOLO-World targets efficient real-time inference
~\cite{minderer2022owlvit,minderer2024owlv2,cheng2024yoloworld}. These systems
are normally compared by average detection accuracy or category transfer.
\method\ instead asks whether a fixed detector remains stable under controlled
rewording of the query.

\textbf{OVD evaluation and robustness:}
COCO and LVIS provide standard detection annotations
~\cite{lin2014coco,gupta2019lvis}, but ordinarily associate each category with
one canonical name. OVDEval studies hallucination and vocabulary ambiguity,
FG-OVD emphasizes fine-grained attributes and parts, and other work studies
visual distribution shifts~\cite{yao2024ovdeval,bianchi2024fgovd,chhipa2024shift}.
These directions are complementary: we hold visual evidence and intended class
fixed and perturb the language interface. The aim is not to replace AP, but to
supplement it with diagnostics that expose prompt-dependent reliability.

\textbf{Prompting, adaptation, and semantic hierarchy:}
Template ensembling and learned context vectors improve transfer in
vision--language models~\cite{radford2021clip,zhou2022coop,zhou2022cocoop};
low-rank prompt adaptation offers another route to efficient specialization
~\cite{zhang2025lowrank}. User-facing studies likewise show that phrasing and
prompt enhancement affect open-set detection~\cite{lin2026xr}. SHiNe builds a
Semantic Hierarchy Nexus offline and substitutes a hierarchy-enriched classifier
without requiring multiple detector passes~\cite{liu2024shine}; hierarchical
consistency is also an emerging OVD objective~\cite{lee2026hierarchical}. Our
benchmark asks a different question: whether improved average accuracy implies
invariance across equivalent surface forms. The matched SHiNe experiment in
Sec.~\ref{sec:shine} shows that these properties are related but not identical.

\section{Benchmark: Lexical Sensitivity in OVD}
\subsection{Detection setup}
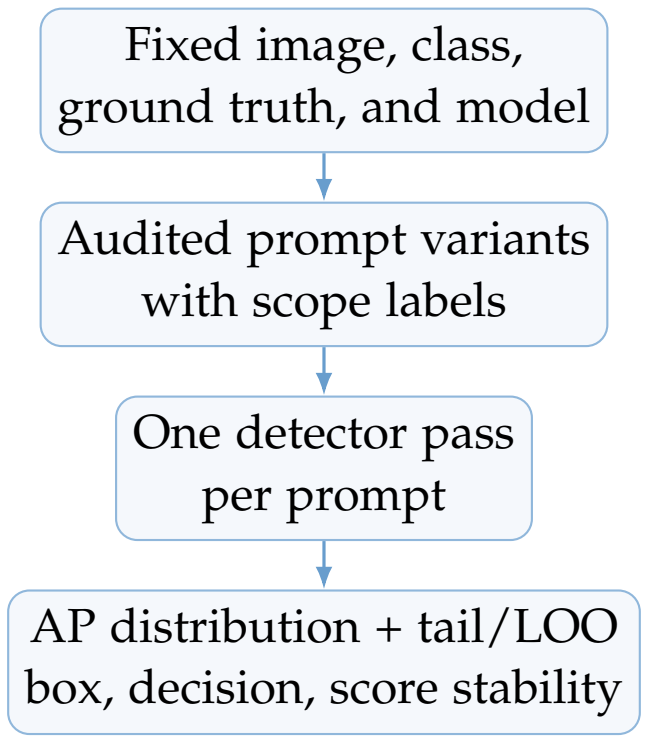
\begin{figure}[t]
\centering
\begin{tikzpicture}[font=\small,node distance=3.2mm,
box/.style={draw=wacvblue!55,rounded corners,fill=wacvblue!5,align=center,inner sep=3.2pt},
arr/.style={-{Latex[length=1.5mm]},semithick,draw=wacvblue!75}]
\node[box] (fixed) {Fixed image, class,\\ground truth, and model};
\node[box,below=of fixed] (prompt) {Audited prompt variants\\with scope labels};
\node[box,below=of prompt] (infer) {One detector pass\\per prompt};
\node[box,below=of infer] (metric) {AP distribution + tail/LOO\\box, decision, score stability};
\draw[arr] (fixed)--(prompt);
\draw[arr] (prompt)--(infer);
\draw[arr] (infer)--(metric);
\end{tikzpicture}
\caption{\method\ isolates the language interface: visual evidence and the
intended class stay fixed while audited prompt realizations vary.}
\label{fig:pipeline}
\end{figure}

Let $I_i\in\mathcal{I}$ be image $i$ and let $p$ be a text prompt. An
open-vocabulary detector is a function
\begin{equation}
D(I_i,p)=\{(b_j,s_j,\ell_j)\}_{j=1}^{M},
\label{eq:detector}
\end{equation}
where $b_j=(x_1,y_1,x_2,y_2)$ is a predicted box, $s_j\in[0,1]$ is confidence,
$\ell_j$ is the predicted text label, and $M$ is the number of returned
detections. For category $c$, the benchmark defines an approved prompt set
$P_c=\{p_1,\ldots,p_K\}$ and ground-truth boxes
$G_i^c=\{g_{i,1}^c,\ldots,g_{i,n_i^c}^c\}$. It evaluates tuples $(I_i,c,p)$ by
comparing $D(I_i,p)$ with $G_i^c$. The central question is whether
$D(I_i,p_a)$ and $D(I_i,p_b)$ behave similarly when $p_a$ and $p_b$ preserve
the same category meaning. Figure~\ref{fig:pipeline} summarizes this controlled
evaluation flow.

\begin{table*}[t]
\centering
\small
\setlength{\tabcolsep}{5.2pt}
\begin{tabularx}{\textwidth}{@{}p{0.9in}Xp{1.65in}p{1.65in}@{}}
\toprule
Scope & Included prompt types & Representative bicycle prompt & Evaluation role \\
\midrule
Equivalent & canonical, template, synonym, plural, paraphrase
  & ``bicycle'', ``bike'', ``bicycles'' & primary lexical robustness \\
Conditional & attribute, context, compositional
  & ``small bicycle'', ``bicycle on the road'' & image-dependent stress test \\
Broad & hypernym, multihop ancestor
  & ``vehicle'', ``mode of transportation'' & taxonomic-scope stress test \\
\bottomrule
\end{tabularx}
\caption{Prompt taxonomy and semantic scope. Only equivalent prompts are
expected to preserve the evaluated ground-truth set and enter the primary
robustness score:}
\label{tab:scope}
\end{table*}

\subsection{Data, prompts, and semantic scope}
\coco\ selects 1,000 COCO 2017 images to cover 30 target classes.
We reserve 200 images for calibration; the primary evaluation uses the other
800, containing 4,136 target instances. \lvis\ contains 1,000 LVIS v1 images, a fixed 30-class set, and 3,897
annotations; its classes are stratified as 11 frequent, 11 common, and 8 rare.
The random seed is 2027. Its first 300 images form an exact nested subset,
allowing a scale check without changing the class definition.

We combine curated names with fixed templates, rule-based paraphrases and
plural inflections. We check WordNet candidates against noun-synset lemmas
and their ancestors~\cite{miller1995wordnet}; each candidate receives a stable
identifier, type and scope before detector inference. The resulting inventory contains 379
\coco\ prompts (193 equivalent, 150 conditional, and 36 broad) and 174
\lvis\ prompts (120 equivalent, 30 conditional, and 24 broad). Every class
has a canonical prompt; the equivalent sets average 6.43 prompts per COCO
class and contain four prompts per LVIS class. Two raters independently assessed all 313 originally equivalent prompts for
scope and naturalness. After adjudication, 187 COCO and 119 LVIS prompts met
both criteria. Supplementary Sec.~1 gives the rules and rater agreement;
the accompanying manifests include every candidate and audit decision.

Table~\ref{tab:scope} makes the semantic distinction explicit. Canonical names,
templates, accepted synonyms, plurals, and direct paraphrases define the
\textbf{equivalent} scope and enter the primary robustness metrics. Attributes,
contexts, and compositional descriptions are \textbf{conditional}: their truth
depends on image content. Hypernyms and multihop ancestors are \textbf{broad}:
they change taxonomic specificity. The latter two scopes are useful diagnostic
queries, but are never pooled with equivalent prompts.

\subsection{Evaluation protocols}
The \textbf{positive-class protocol} queries a class only on selected images
containing that class. It isolates lexical variation conditioned on object
presence and is run for all four detectors. A secondary \textbf{canonical
all-class protocol} queries one audited canonical name per class on every COCO
image for Grounding DINO and OWL-ViT. It includes absent-category queries and
therefore serves as a false-positive stress test. To isolate absence from
paraphrase variation, it does not multiply absent-class queries by every
wording. A separate 100-image COCO pilot tests absent-category wordings with
OWL-ViT (Supplementary Sec.~4). Primary robustness scores use only the
positive protocol.

\section{Prompt-Robustness Metrics}
\subsection{Prompt-distribution accuracy}
For class task $c$ and approved equivalent prompt $p\in P_c$, let $A_{c,p}$ be
average precision over the selected images at intersection-over-union (IoU)
0.5. IoU is the area of intersection of two boxes divided by their union, and
AP$_{50}$ is the area under the ranked precision--recall curve when IoU is at
least 0.5. We define
\begin{align}
\mu_c&=\frac{1}{|P_c|}\sum_{p\in P_c}A_{c,p}, &
W_c&=\min_{p\in P_c}A_{c,p},\\
B_c&=\max_{p\in P_c}A_{c,p}, &
G_c&=B_c-W_c.
\label{eq:distribution}
\end{align}
These are mean-prompt AP, worst-prompt AP, best-prompt oracle, and prompt gap.
The oracle selects a prompt using test labels and is diagnostic, not deployable.
A detector can have high mean AP but low worst-prompt AP if a few ordinary
variants fail badly; the gap expresses this spread directly.

Because one difficult prompt can dominate a minimum, we also report the
empirical 10th percentile (P10) and CVaR$_{20}$, the arithmetic mean of the
lowest $\lceil0.2|P_c|\rceil$ prompt scores. With at most five prompts,
including all LVIS classes, CVaR$_{20}$ equals the minimum. Two leave-one-prompt-out (LOO)
statistics quantify the dependence on a single prompt:
\begin{align}
I^W_c &= \max_j\min_{p\ne j} A_{c,p}-W_c,\\
I^G_c &= G_c-\min_j\left(\max_{p\ne j}A_{c,p}-\min_{p\ne j}A_{c,p}\right).
\label{eq:loo}
\end{align}
$I^W_c$ is the largest rise in worst-prompt AP after removing one prompt, while
$I^G_c$ is the largest reduction in the gap. Dataset summaries macro-average
the 30 class tasks. Paired confidence intervals use 10,000 class-block bootstrap
resamples and seed 2027.

\subsection{Spatial, decision, and score stability}
AP does not reveal whether different prompts localize the same region. For each
ground-truth instance $i$ and prompt $p$, we therefore match detections by
Hungarian one-to-one assignment at IoU$\ge0.5$. The one-to-one constraint
prevents two predictions from claiming the same object. Let $z_{i,p}=1$ when a
match exists and 0 otherwise, and let
$b_{i,p}$ and $s_{i,p}$ be its box and confidence when matched.

For each instance $i$, let $\langle\cdot\rangle$ denote the average over
its equivalent prompts $p\in P$ or unordered pairs $p<q$. Write
$h=z_{i,p}z_{i,q}$ and $\rho=\operatorname{IoU}(b_{i,p},b_{i,q})$ when
both prompts detect the instance; set $\rho=0$ otherwise. A missed prompt
has zero-filled confidence $\tilde s_{i,p}=0$; otherwise $\tilde s_{i,p}=s_{i,p}$.
\begin{align*}
\mathrm{DR}_i &= \langle z_{i,p}\rangle_p,\\
\mathrm{Agr}_i &= \langle\mathbf{1}[z_{i,p}=z_{i,q}]\rangle_{p<q},\\
\mathrm{Jac}_i &= \langle h\rangle_{p<q:\,z_{i,p}+z_{i,q}>0}.
\end{align*}
DR is detection rate; agreement includes joint misses, while Detection
Jaccard conditions on at least one detection. For boxes and scores,
\begin{align*}
\mathrm{Box}_{0,i} &= \langle \rho\rangle_{p<q},\\
\mathrm{Box}_{m,i} &= \langle \rho\rangle_{p<q:\,h=1},\\
\mathrm{Var}_{0,i} &= \operatorname{Var}_{p\in P}(\tilde s_{i,p}),\\
\mathrm{Var}_{m,i} &= \operatorname{Var}_{p:\,z_{i,p}=1}(s_{i,p}).
\end{align*}
Box$_0$ assigns zero to either miss; Box$_m$ uses joint detections.
Var$_0$ includes misses; Var$_m$ requires at least two matched scores.
For each metric we average defined values over instances within each class,
then equally over classes with a defined value (Supplementary Sec.~1).

\subsection{Reading and reusing the benchmark}
We recommend starting with mean AP, gap and prompt count for accuracy;
detection rate and Jaccard for missed objects; and matched box IoU for
localization. Table~\ref{tab:metric-reading} summarizes common combinations. For example,
high agreement is not automatically reassuring: if detection rate is
low, two prompts may agree because both miss the object. Similarly, high
Box$_m$ with much lower Box$_0$ means that the model localizes consistently
when it fires, but whether it fires remains wording-dependent. Tail scores help when enough prompts are available; LOO measures the
influence of a single wording. These combinations make the benchmark
useful for diagnosing a deployed detector rather than merely assigning it a
single robustness rank.

\begin{table}[tbp]
\centering
\small
\setlength{\tabcolsep}{3.2pt}
\begin{tabular}{p{.40\linewidth}p{.50\linewidth}}
\toprule
Observed pattern & Likely interpretation \\
\midrule
High mean; low P10/CVaR & one or more wordings fail \\
High agreement; low DR & prompts frequently agree by missing \\
High Box$_m$; low Box$_0$ & boxes are stable conditional on detection \\
High Box$_m$; high Var$_0$ & location is stable but confidence changes \\
Broad $\ll$ equivalent & hierarchy/scope stress, not equivalent drift \\
Aggregator $<$ oracle & prompt-selection headroom remains \\
\bottomrule
\end{tabular}
\caption{Metric reading guide. Complementary statistics localize different failure modes; no
single row is treated as a complete robustness score.}
\label{tab:metric-reading}
\end{table}

For reuse, an extension should preserve four invariants. First, image IDs,
class IDs, annotations, model checkpoint, and detector thresholds must remain
fixed across prompts. Second, every prompt needs a stable identifier, type,
scope, and auditable acceptance decision; rejected candidates should remain in
the record so the final vocabulary is reproducible. Third, primary robustness
must be computed only within an explicitly equivalent scope, while conditional,
broad, and absent-class tests remain separate. Fourth, results should be
macro-averaged over class tasks and statistical resampling should preserve all
prompts and compared methods within a sampled class. These constraints allow
new datasets, detectors, languages, or prompt generators to be added without
changing what the reported quantities mean.

\section{Prompt Aggregation}
Prompt disagreement can also be used as an inference-time signal. We compare
four training-free aggregators on identical saved detections. Source boxes below
0.01 are discarded, at most 20 boxes per prompt are retained, boxes are linked
at IoU 0.5, each connected component receives at most one vote per prompt, and
final non-maximum suppression (NMS) uses IoU 0.5.

\textbf{Simple baselines:} Max-score retains the highest-confidence member of
each component. Mean weighted-box fusion (Mean-WBF) outputs a score-weighted
coordinate average and uses the mean vote score. They isolate the value of
multi-prompt ensembling without a learned reliability model.

\textbf{Prompt Consensus:} For component $k$, let $m_k$ be mean confidence and
$u_k$ the fraction of source prompts represented. \pc\ assigns
\begin{equation}
s_k=\alpha m_k+(1-\alpha)u_k,
\label{eq:consensus}
\end{equation}
with $\alpha=0.5$, keeps components meeting a prompt-support threshold, outputs
a score-weighted box, and applies final NMS. The benchmark comparison uses 25\%
support. The separate four-prompt Detic study pre-specifies 50\% support and
reports 25--100\% sensitivity. The method remains intentionally simple:
agreement across wording is treated as a lightweight reliability signal, not a
claim that every prompt deserves equal trust.

\textbf{Exploratory LSF:} Lexical Stability Fusion (LSF) estimates
model-specific prompt-type reliability from unlabeled box agreement on a frozen
20\% COCO calibration split. For component $k$, it combines a
reliability-weighted mean score $m_k$, weighted support $u_k$, median pairwise
IoU $c_k$, and weighted median absolute score deviation $d_k$:
\begin{equation}
s_k=m_k u_k^{\gamma}c_k^{\delta}\exp(-\beta d_k).
\label{eq:lsf}
\end{equation}
Its output is an observed reliability-and-score-weighted medoid, avoiding an
unsupported averaged coordinate. A 27-point grid on the calibration split
selects $(\beta,\gamma,\delta)=(0.5,0.5,0)$ before held-out COCO and all LVIS
evaluation. The optimum barely exceeds its runner-up and learned type weights
are nearly uniform, so LSF is an exploratory ablation rather than a primary
contribution.

\section{Experiments}
\subsection{Experimental setup}
We evaluate public checkpoints for IDEA-Research Grounding-DINO-Tiny, Google
OWL-ViT Base Patch32, Google OWLv2 Base Patch16 Ensemble, and YOLOv8-L
World-v2. Score and text thresholds are 0.001, NMS IoU is 0.5, and at most 100
detections are retained. To separate lexical sensitivity from operating-point
choice, we report a prespecified nine-point score sweep from 0.001 to 0.5. Selected
image manifests, prompt audit decisions, model identifiers, environment
snapshots, and SHA-256 fingerprints are retained; no inference errors occur in
the reported cells. Hardware, software, runtime, and cost details are separated
into Supplementary Sec.~5.

\subsection{Equivalent-prompt sensitivity}
Table~\ref{tab:main} presents the primary results. Average AP conceals large
worst-prompt failures, while the four-family comparison also reveals important
architectural differences. Grounding DINO has a 23.4-point prompt gap on
\coco\ and a 29.9-point gap on \lvis. OWL-ViT gaps are 19.8 and 12.2 points.
OWLv2 is markedly more stable, with gaps near 5.2 points on both datasets,
whereas YOLO-World-v2 is stable on COCO (6.6 points) but less so on LVIS (13.0
points). Strong mean accuracy and lexical stability are therefore distinct
properties: Grounding DINO and YOLO-World-v2 have similar COCO means, yet their
gaps differ by 16.8 points. Restricting the prompts to the 306 retained
after adjudication preserves the detector ordering by gap on both datasets
(Supplementary Sec.~2).

With four LVIS prompts, P10 interpolates between the two lowest scores.
On LVIS, Grounding DINO's mean, P10, and CVaR$_{20}$ are 51.6, 38.8, and 35.0,
while OWLv2 reaches 71.6, 69.5, and 68.7. Removing one prompt can raise
Grounding DINO's LVIS worst score by 12.8 points and shrink its gap by 15.0
points. We therefore recommend reporting a minimum together with P10/CVaR and
LOO influence rather than treating any one statistic as sufficient.

\begin{table*}[t]
\centering
\small
\setlength{\tabcolsep}{5.4pt}
\begin{tabular}{llrrrrrr}
\toprule
Dataset & Model & Mean & Worst & P10 & CVaR$_{20}$ & Best & Gap \\
\midrule
\multirow{4}{*}{\coco}
 & Grounding DINO-T & 75.22 & 60.36 & 66.58 & 65.70 & 83.78 & 23.42 \\
 & OWL-ViT-B/32 & 62.64 & 47.89 & 54.59 & 53.79 & 67.67 & 19.79 \\
 & OWLv2-B/16-Ens. & \best{78.67} & \best{74.81} & \best{76.92} & \best{76.61} & 80.16 & \best{5.35} \\
 & YOLO-World-v2-L & 75.77 & 72.28 & 73.33 & 73.18 & 78.89 & 6.61 \\
\midrule
\multirow{4}{*}{\lvis}
 & Grounding DINO-T & 51.60 & 34.96 & 38.79 & 34.96 & 64.87 & 29.91 \\
 & OWL-ViT-B/32 & 39.83 & 32.28 & 34.42 & 32.28 & 44.51 & 12.23 \\
 & OWLv2-B/16-Ens. & \best{71.59} & \best{68.65} & \best{69.52} & \best{68.65} & 73.78 & \best{5.13} \\
 & YOLO-World-v2-L & 49.66 & 42.66 & 44.45 & 42.66 & 55.64 & 12.98 \\
\bottomrule
\end{tabular}
\caption{Equivalent-prompt results under the positive protocol
(\apfifty$\times100$; 800 held-out COCO and 1,000 LVIS images).
Best is a non-deployable prompt oracle; Gap$=$Best$-$Worst.}
\label{tab:main}
\end{table*}

\subsection{Detection, localization, and confidence stability}
\begin{figure}[tbp]
\centering
\begin{minipage}[t]{.485\columnwidth}\centering
\includegraphics[width=\linewidth]{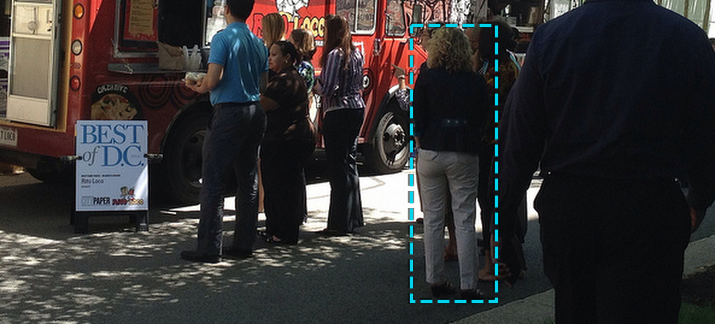}\par
{\footnotesize \emph{human} (0.087)}
\end{minipage}\hfill
\begin{minipage}[t]{.485\columnwidth}\centering
\includegraphics[width=\linewidth]{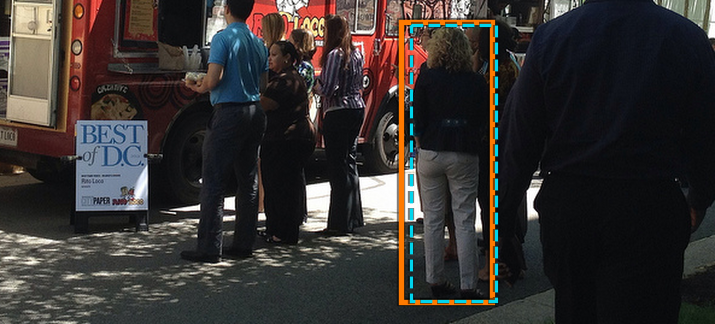}\par
{\footnotesize \emph{person} (0.199)}
\end{minipage}
\caption{OWL-ViT proposes the same box for the person at the food truck
under both prompts. Only ``person'' clears the 0.10 threshold (orange:
matched detection; dashed cyan: ground truth).}
\label{fig:foodtruck}
\end{figure}

Table~\ref{tab:stability} reports the metrics that AP alone cannot reveal.
OWLv2 combines high detection rate with near-perfect matched localization
consistency. Grounding DINO also has high detection rate, but lower Box$_0$,
consistent with prompt-dependent misses or localization changes. Detection
agreement can remain high when prompt pairs jointly miss an instance, which is
why it is reported with detection rate, Jaccard, and zero-inclusive box
coverage. The low confidence variance of OWLv2 shows that its stability is not
only spatial. Conversely, YOLO-World-v2 on LVIS has matched box consistency of
93.9\% but Box$_0$ of 57.9\%: when two prompts find an object, their boxes agree,
but prompt-dependent misses remain common. Figure~\ref{fig:foodtruck} shows a threshold crossing for OWL-ViT on COCO.

\begin{table*}[t]
\centering
\small
\setlength{\tabcolsep}{4.7pt}
\begin{tabular}{llrrrrrrr}
\toprule
Dataset & Model & DR & Agr. & Jac. & Box$_0$ & Box$_m$ & Var$_0$ & Var$_m$ \\
\midrule
\multirow{4}{*}{\coco}
 & Grounding DINO-T & 93.88 & 96.34 & 94.51 & 84.68 & 91.09 & 16.94 & 17.05 \\
 & OWL-ViT-B/32 & 86.52 & 99.23 & 98.65 & 85.24 & 98.97 & 17.53 & 19.37 \\
 & OWLv2-B/16-Ens. & 94.13 & 99.76 & 99.50 & 93.69 & 99.66 & 2.47 & 2.59 \\
 & YOLO-World-v2-L & 87.80 & 96.11 & 92.27 & 82.46 & 95.41 & 16.53 & 18.23 \\
\midrule
\multirow{4}{*}{\lvis}
 & Grounding DINO-T & 83.16 & 92.05 & 85.70 & 72.14 & 90.25 & 12.90 & 13.24 \\
 & OWL-ViT-B/32 & 71.49 & 98.61 & 96.54 & 69.62 & 98.41 & 4.66 & 5.99 \\
 & OWLv2-B/16-Ens. & 90.94 & 99.69 & 99.43 & 90.69 & 99.90 & 1.62 & 1.83 \\
 & YOLO-World-v2-L & 66.34 & 89.88 & 73.87 & 57.85 & 93.89 & 8.10 & 9.49 \\
\bottomrule
\end{tabular}
\caption{Instance stability on original equivalent prompts. DR, agreement,
Jaccard and box IoUs are percentages; both confidence variances are multiplied
by $10^3$. Matched IoU and variance use 29 classes for LVIS YOLO-World-v2
(no joint match for \emph{die}); other entries use 30.}

\label{tab:stability}
\end{table*}

\subsection{What the prompt taxonomy reveals}
Table~\ref{tab:semantic} scores each scope against the original child-class
boxes. For \emph{small bicycle}, all bicycles count; no image-level label
indicates which are small. Conditional and broad
prompts are often harder, but these decreases must not be interpreted as
meaning-preserving lexical failures. For example, ``vehicle'' can correctly
refer to a bus or motorcycle when the evaluated child class is ``bicycle'';
counting those boxes as bicycle false positives conflates semantic scope with
detector error. Broad AP does not score the union of descendant classes, and conditional
AP cannot establish whether a detector follows an attribute.

The four-model pattern also rules out a simple ``newer is always more robust''
story. OWLv2 is stable across equivalent prompts but drops sharply for broad
terms, which is expected because the query itself changed. YOLO-World-v2 is
stable on COCO yet less stable on LVIS. Architecture, vocabulary, dataset, and
semantic scope interact; one pooled prompt average would conceal that structure.

\begin{table}[t]
\centering
\small
\footnotesize
\setlength{\tabcolsep}{2.0pt}
\begin{tabular}{llrrr}
\toprule
Dataset & Model & Equivalent & Conditional & Broad \\
\midrule
\multirow{4}{*}{\coco}
 & GDINO-T & 75.22 & 74.12 & 56.36 \\
 & OWL-ViT & 62.64 & 56.07 & 29.00 \\
 & OWLv2 & 78.67 & 79.18 & 29.91 \\
 & YOLO-W-v2 & 75.77 & 74.28 & 52.77 \\
\midrule
\multirow{4}{*}{\lvis}
 & GDINO-T & 51.60 & 47.49 & 35.30 \\
 & OWL-ViT & 39.83 & 43.05 & 22.05 \\
 & OWLv2 & 71.59 & 71.21 & 31.04 \\
 & YOLO-W-v2 & 49.66 & 39.98 & 24.03 \\
\bottomrule
\end{tabular}
\caption{Mean \apfifty$\times100$ by semantic scope. Conditional and broad
prompts are scored against child-class labels and excluded from the
equivalent-prompt mean.}
\label{tab:semantic}
\end{table}

\begin{table*}[t]
\centering
\small
\setlength{\tabcolsep}{4.4pt}
\begin{tabular}{llrrrrrr}
\toprule
Dataset & Model & Mean & Oracle & Max-score & Mean-WBF & Consensus & LSF \\
\midrule
\multirow{4}{*}{\coco}
 & Grounding DINO-T & 75.22 & 83.78 & 78.10 & 75.89 & 78.16 & \best{79.93} \\
 & OWL-ViT-B/32 & 62.64 & 67.67 & 66.38 & 66.68 & \best{67.02} & 66.99 \\
 & OWLv2-B/16-Ens. & 78.67 & 80.16 & \best{79.19} & 79.13 & 79.13 & 79.02 \\
 & YOLO-World-v2-L & 75.77 & 78.89 & 75.83 & 76.01 & 75.43 & \best{76.25} \\
\midrule
\multirow{4}{*}{\lvis}
 & Grounding DINO-T & 51.60 & 64.87 & 55.72 & 54.47 & 56.60 & \best{57.47} \\
 & OWL-ViT-B/32 & 39.83 & 44.51 & 42.56 & 42.17 & 43.26 & \best{43.39} \\
 & OWLv2-B/16-Ens. & 71.59 & 73.78 & \best{69.81} & 69.44 & 69.15 & 69.37 \\
 & YOLO-World-v2-L & 49.66 & 55.64 & 48.51 & 49.07 & \best{50.05} & 49.62 \\
\bottomrule
\end{tabular}
\caption{Aggregation comparison (\apfifty$\times100$). Mean is the average
single-prompt baseline; Oracle selects the best prompt using test labels. Bold
marks the best deployable method in each row, not a universal winner.}
\label{tab:aggregation}
\end{table*}

\subsection{Class-level failure patterns}
Dataset averages conceal a second kind of structure: the most sensitive
classes differ across detector families. Supplementary Sec.~6 reports positive
image and instance counts alongside each gap and its 95\% image-bootstrap
interval; rankings on sparse classes are uncertain. On COCO, Grounding DINO's five
largest equivalent-prompt gaps occur for traffic light, car, sheep, person, and
cow (37.0--48.8 points). OWL-ViT instead peaks on television and dog, followed
by toilet, zebra, and airplane (30.6--47.7 points). OWLv2 has a much shorter
tail, although laptop and traffic light still reach 24.2 and 20.0 points.
YOLO-World-v2 is also comparatively stable, with its largest gap, for cell
phone, at 20.9 points. Thus a model can have a small dataset-level gap while
retaining isolated classes for which wording is consequential.

LVIS exposes a related pattern with less familiar names. Grounding DINO is
most sensitive for radio receiver, stepladder, traffic light, pillow, and
kitchen table; its class-level gaps range from 50.1 to 65.0 points. OWL-ViT's
largest gap is 47.1 points for kettle, while YOLO-World-v2 reaches 73.3 points
for stepladder. OWLv2 remains the most stable family overall, but radio receiver
still has a 20.1-point gap. These are not uniformly small objects, rare nouns,
or members of one visual domain. The repeated appearance of traffic light,
radio receiver, and stepladder across some models suggests difficult visual or
lexical concepts, whereas the non-overlapping failures show that architecture
also matters.

These rankings describe the sampled images. For example, the 73.3-point
stepladder gap for YOLO-World-v2 comes from just five positive images.
Broad family names remain excluded from the equivalent-prompt gap.

\begin{table*}[t]
\centering
\small
\setlength{\tabcolsep}{4.5pt}
\begin{tabular}{lrrrrl}
\toprule
\multicolumn{6}{l}{\textbf{(a) Prompt-wise robustness statistics}} \\
\midrule
Method & Passes & Mean AP$_{50}$ & Worst AP$_{50}$ & Gap & Paired interpretation \\
\midrule
Detic statistics & 4 & 58.95 & 54.28 & 8.30 & prompt distribution \\
SHiNe fixed nexus & 4 & 59.31 & 55.70 & 6.67 & $\Delta_{\rm mean}=+0.36$ \\
\bottomrule
\end{tabular}

\vspace{2pt}

\begin{tabular}{llrrl}
\toprule
\multicolumn{5}{l}{\textbf{(b) Single-output official and aggregation results}} \\
\midrule
Study & Method & Passes & Output AP$_{50}$ & Paired interpretation \\
\midrule
Official & Detic & 1 & 60.54 & reference \\
Official & Detic+SHiNe & 1 & 60.33 & $\Delta=-0.22$, CI $[-0.93,0.38]$ \\
4-prompt & Max-score & 4 & 60.20 & aggregation \\
4-prompt & Mean-WBF & 4 & \best{61.04} & aggregation \\
4-prompt & Consensus ($u\ge.50$) & 4 & 59.97 & $+1.01$ vs. prompt mean \\
\bottomrule
\end{tabular}
\caption{Detic/SHiNe on the matched \lvis\ set. Official methods use one pass;
panel (a) reports prompt-wise statistics over canonical, template, plural, and
paraphrase passes, while panel (b) reports the single output from official or
aggregated methods.
Both panels report all 30 classes; paired intervals here resample those
30 class blocks, not just the eight hierarchy-active classes.}
\label{tab:shine}
\end{table*}

\subsection{Aggregation helps selectively}
Table~\ref{tab:aggregation} places every aggregation rule beside the mean
single-prompt baseline and best-prompt oracle. The oracle is useful for
measuring remaining headroom but cannot be deployed because it uses test labels.
No deployable method dominates all eight model--dataset cells. On COCO, \pc\
improves Grounding DINO by 2.94 points and OWL-ViT by 4.38 points over the mean,
while changing OWLv2 and YOLO-World-v2 only modestly. On LVIS, it improves
Grounding DINO and OWL-ViT by 5.00 and 3.43 points, but reduces OWLv2 by 2.44
points. LSF gives the strongest Grounding DINO point estimates but is not a
universal winner either. Prompt dropout at 50\% and 75\% retention preserves the
same qualitative ordering (Supplementary Sec.~3).

We treat Prompt Consensus as a diagnostic aggregation baseline. It can recover
performance lost to an unlucky wording, particularly for less stable models,
but it does not remove the underlying distribution or reach the best-prompt
oracle. We therefore base mitigation claims on the complete, semantically
scoped model--dataset matrix in Table~\ref{tab:aggregation}, not on a single
favorable operating point.

\subsection{Controlled comparison with SHiNe}
\label{sec:shine}
SHiNe is especially relevant because it enriches an OVD classifier with a
semantic hierarchy while keeping inference to one detector pass. We evaluate
official Detic and Detic+SHiNe on the same \lvis\ images, classes, checkpoint,
and detector configuration. The official one-pass classifiers reach 60.54 and
60.33 \apfifty; the paired difference is $-0.22$ points with 95\% CI
$[-0.93,0.38]$. Thus SHiNe does not significantly change accuracy on this
restricted matched set.

We then hold the SHiNe nexus construction fixed while varying the leaf
realization among canonical, template, plural, and paraphrase forms;
Table~\ref{tab:shine} summarizes the resulting comparisons. The
four-prompt Detic distribution has mean 58.95, worst 54.28, and gap 8.30; the
fixed-nexus variant narrows the point-estimate gap to 6.67. Hierarchy enrichment
is active for eight classes; the other 22 serve as controls (Supplementary Sec.~5). At 50\% support,
\pc\ reaches 59.97, a 1.01-point improvement over the Detic prompt mean (CI
$[0.12,1.78]$, $p=.018$), but it is not significantly better than official
Detic or official SHiNe. Mean-WBF has the highest point estimate, 61.04, yet its
advantage over \pc\ is also not established. Semantic hierarchy and prompt
aggregation can therefore reduce selected failures, but neither guarantees
lexical invariance.

\subsection{Threshold and scale checks}
AP is computed from ranked detections, so fixed thresholds mainly change the
candidate pool and threshold-dependent precision, recall, and false positives.
We sweep 0.001, 0.005, 0.01, 0.025, 0.05, 0.1, 0.2, 0.3, and 0.5 for every
model and dataset. The low-threshold AP optimum is stable for Grounding DINO and
OWLv2, while YOLO-World-v2 loses recall sooner. We retain a common 0.001
detector threshold rather than selecting a test-specific operating point. The
complete sweep is in Supplementary Sec.~4.

Expanding LVIS from its 300-image prefix to 1,000 images changes mean-prompt
\apfifty\ by only $-0.24$ to $-1.46$ points and preserves the four-model
ordering. The additional images therefore strengthen the estimate without
reversing the smaller-sample pattern.

\subsection{Uncertainty and prompt-set dependence}
Because the same 30 classes are evaluated under every model and method, we use
paired class-block rather than independent resampling. The resulting intervals
show that mean accuracy and robustness can separate. On COCO, OWLv2 exceeds
Grounding DINO by 3.46 mean-AP points (95\% CI $[1.22,5.74]$), but by 14.44
worst-prompt points (CI $[10.94,18.04]$); its prompt gap is 18.07 points smaller
(CI $[14.46,21.84]$). On LVIS, the corresponding differences are 19.99 in mean
AP, 33.69 in worst-prompt AP, and 24.78 in gap, with all paired intervals
excluding zero. By contrast, OWLv2 and YOLO-World-v2 have statistically
distinguishable COCO means, but their worst-prompt and gap intervals overlap
zero. It would therefore be inaccurate to rank them as different in lexical
robustness from those point estimates alone.

Prompt dropout asks whether an aggregation conclusion depends on retaining the
entire curated set. We repeatedly retain 50\% or 75\% of prompts and recompute
each method from the saved predictions. On LVIS, Grounding DINO LSF changes
from 57.47 AP with all prompts to 56.52 at 75\% and 54.52 at 50\%; Prompt
Consensus follows the same gradual pattern (56.60, 56.31, and 54.38). For the
more stable OWLv2, all corresponding changes are below one AP point at 50\%
retention. Supplementary Sec.~3 gives the dropout results. Supplementary Sec.~2
separately controls for prompt count (two per class) and reports
AP$_{75}$ gaps and class-level metric correlations.

\section{Discussion}
\textbf{Evaluation implications:}
The language interface is part of the detection system. Mean AP alone can hide
user-facing brittleness,
but a minimum alone can also be controlled by one poor prompt. Mean, worst,
P10/CVaR, gap, and LOO answer complementary questions. Instance matching adds a
second axis: a detector can preserve AP while shifting boxes or confidence, or
it can localize consistently whenever successful while missing under particular
wordings. The metric suite is intended to expose these cases, not collapse them
into one scalar.

\textbf{Semantic implications:}
Equivalent, conditional, and broad prompts should not be mixed. The scope
labels make the interpretation precise: broad queries change which objects
the words describe, so their child-class AP has a different interpretation
from instability among equivalent names. The canonical all-class protocol is
similarly separate from prompt robustness; it measures whether an arbitrary
absent-category query produces detections, not how equivalent prompts vary.

\textbf{Method implications:}
Prompt agreement remains a useful lightweight reliability signal. Boxes
supported by several equivalent wordings are often more reliable than
prompt-specific boxes, especially for Grounding DINO and OWL-ViT. However,
multi-prompt methods require $K$ detector passes for $K$ prompts, whereas the
official SHiNe classifier uses one. The best-prompt oracle remains above every
deployable aggregator, and model-dependent degradations rule out a universal
mitigation claim. The practical role of aggregation is therefore to reduce the
risk of an unlucky prompt when its additional inference cost is acceptable.

\section{Limitations}
The benchmark is controlled but not exhaustive. Both datasets use one
deterministic image seed and 30 classes; \lvis\ is not the full 1,203-class LVIS
benchmark. Only two models enter the canonical all-class protocol and only OWL-ViT
the absent-wording pilot. The human audit covers English prompts initially
labeled equivalent; it does not verify attributes in individual images.
Conditional and broad prompts remain diagnostics. AP$_{50}$ is the primary robustness measure,
with AP$_{75}$ reported as a localization check. Prompt dropout resamples
saved prompt outputs rather than stochastic detector seeds. Only eight classes received fixed-nexus SHiNe enrichment. Broader class
coverage, larger checkpoints, multiple image seeds, multilingual prompts, human-authored queries,
and explicit deployment-cost studies remain important extensions.

\section{Conclusion}
We introduced \method, a controlled benchmark and metric suite for measuring
lexical sensitivity in open-vocabulary object detection. The central message is
that the language interface of a detector is part of the evaluation problem,
not merely a convenient way to name categories. Across four detector families
on the held-out COCO and LVIS evaluations, equivalent wording changes produce gaps as large as
29.9 \apfifty\ points and measurable changes in detection state, localization,
and confidence. Tail and leave-one-out statistics prevent a single prompt from
silently controlling the conclusion, while one-to-one matching makes the
analysis valid when several instances of a class are present.

The results also clarify the role of mitigation. Prompt aggregation and
hierarchy-aware classifiers can reduce sensitivity in selected settings, but
neither removes it consistently and both accuracy and computational cost must
be considered. Reliable open-vocabulary perception should therefore report
performance over semantically scoped prompt distributions, disclose the tail
and best-prompt oracle, and treat prompt selection as part of the evaluated
system. Robustness to linguistic variation complements, rather than replaces,
standard detection accuracy.

\clearpage
\appendix
\setcounter{section}{0}
\setcounter{subsection}{0}
\setcounter{table}{0}
\setcounter{figure}{0}
\setcounter{equation}{0}
\renewcommand{\thesection}{S\arabic{section}}
\renewcommand{\thetable}{S\arabic{table}}
\renewcommand{\thefigure}{S\arabic{figure}}
\renewcommand{\theequation}{S\arabic{equation}}
\renewcommand{\theHsection}{supp.\arabic{section}}
\renewcommand{\theHsubsection}{supp.\arabic{section}.\arabic{subsection}}
\renewcommand{\theHtable}{supp.\arabic{table}}
\renewcommand{\theHfigure}{supp.\arabic{figure}}
\renewcommand{\theHequation}{supp.\arabic{equation}}

\section*{Supplementary Material}
The accompanying ZIP (\href{https://www.dropbox.com/scl/fi/kjtkwrjmv06bvfg9iettj/02112_supp.zip?rlkey=obxvgdb4o023rs0i2xymu98qg&st=e2zc1e1x&dl=0}{link here}) includes the full candidate inventories, human
scope and naturalness decisions, frozen image lists, and numerical tables.
The human audit covers originally equivalent prompts only.

\section{Protocol and prompt construction}
\paragraph{Image sampling.}
We select COCO images to cover the 30 target classes, breaking ties
deterministically (seed 2027), then favoring remaining images with a selected
category. We assign the first 200 IDs in seeded hash order to calibration and
the remaining 800 to evaluation. The 1,000-image LVIS selection extends the
frozen 300-image prefix without changing its class set. The primary evaluations
contain 10,021 and 4,704 equivalent image--class--prompt queries per model,
respectively. The 100-image negative pilot uses a separately hashed ordering
of the 800 evaluation IDs and has no overlap with calibration.

\paragraph{How the words were generated.}
COCO starts with curated names, synonyms, plurals, three template forms
(“a/an [class]”, “a photo of a/an [class]”, “a visible [class]”), and
attribute, context and parent-category phrases. Deterministic rules add
“the [class] in the image” and “a clearly visible [class] in the scene”.
LVIS uses the class name with underscores replaced by spaces, the photographic
template, its plural and the referential paraphrase. The generator passes
the full noun phrase to the English inflection routine, including for
multiword names; the manifest records the actual output for every class.
There is no language-model paraphrasing step. The audited vocabulary, not
regeneration with a different inflector or WordNet release, defines this study.

Curated COCO synonyms are looked up in a versioned class--string decision table.
For WordNet candidates the audit tests synonyms against noun-synset lemmas,
immediate parent terms against hypernym lemmas, and two-hop terms against
lemmas of a parent's hypernyms. LVIS checks its annotated synset;
the COCO WordNet check excludes a proposed relation if it holds for only
some noun senses of an ambiguous class. Of 441 COCO candidates, 45 curated
synonym candidates and 17 two-hop ancestors were excluded; six two-hop terms
were retained. LVIS excluded five proposed ancestors, leaving 174 of 179
candidates. These exclusions are distinct from subsequent human review.
Some legacy labels encode different conventions for similar strings:
COCO's “a visible bicycle” is a template and “visible bicycle” an attribute.
Scope labels identify evaluation conditions; they are not image annotations.

\paragraph{Human review.}
Two raters independently judged each originally equivalent prompt against its
class definition, without consulting detector outputs. They chose one scope
(equivalent, conditional, broad, ambiguous or invalid) and separately rated
naturalness (natural, awkward or unsure). The archive includes the rating
instructions and class definitions. The adjudicated subset
requires both “equivalent” and “natural”. Five COCO mouse strings were judged
ambiguous between device and animal; “motor cycle” (COCO) and “a photo of a
shampoo” (LVIS) were judged equivalent but awkward. The original inventory
remains available. Of 313 prompts, 219 were called equivalent and natural
by both initial raters. Scope agreement was 70.6\%, close to the 70.7\%
expected from the raters' marginal frequencies; naturalness agreement was
90.1\% versus 89.5\% expected. Unweighted Cohen's $\kappa$ was
$-0.004$ for scope and $0.055$ for naturalness. Thus the final 306 retained prompts should not
be interpreted as near-unanimous linguistic judgments. Ratings concern
English class references, not attribute truth in particular images.

\subsection{Stability aggregation and edge cases}
First compute each stability metric for an individual ground-truth object.
For detection Jaccard remove prompt pairs with two misses; for matched box IoU
retain only pairs with two detections. Exclude an instance from a conditional
metric when its denominator is zero; omit a class only when no finite instance
remains. Thus conditional metrics weight eligible instances equally within
class. Matched confidence variance needs two detected scores and uses a
population variance. The supplied stability tables give the number of finite
instances and classes behind each mean.

\subsection{Statistical-test conventions}
The paired class-block bootstrap samples the 30 class tasks with replacement
while preserving all prompts and compared methods within each sampled class.
Reported 95\% intervals use 10,000 resamples and seed 2027. The two-sided
sign-flip test independently reverses paired class differences; its $p$-value
is the fraction of random absolute mean differences at least as large as the
observed absolute mean difference, with a plus-one correction.

\section{Prompt inventory and localization controls}
Table~\ref{tab:controls} compares the original positive-protocol gaps with
two human filters, a stricter matching IoU, and fixed-size prompt samples.
For each class we draw two distinct original prompts from the allowed types
and average the class gaps across 200 repetitions. “No template” excludes
the entire template category; on COCO it includes article and visible forms
as well as the photographic form. All 30 classes remain eligible in each
of the four sampling conditions. Differences between conditions depend on
the strings available, not just on prompt count. Repetitions reflect draws
of prompts and do not constitute image-level confidence intervals.

Size-matched enumeration gives a second control for the human filter:
enumerate subsets of the original class inventory of the same size as
the adjudicated set. The observed gap minus the average gap over size-matched
subsets ranges from $-0.29$ to $+0.55$ AP points across the eight cells.
The stricter unanimous filter retains only one prompt in one COCO class;
its full-set range is zero, but that class is ineligible for a two-prompt
comparison. The class definition of LVIS \emph{kettle} was flagged for its
cooking-pot sense and retained. The companion also gives the paired
29-class sensitivity analysis omitting that class.

\begin{table*}[t]\centering\small
\setlength{\tabcolsep}{4.0pt}
\begin{tabular}{llrrrrrrrr}\toprule
 & & \multicolumn{3}{c}{AP$_{50}$ gap by prompt inventory} & AP$_{75}$ & \multicolumn{4}{c}{AP$_{50}$ gap with two prompts per class}\\
Dataset & Model & Original & Adjud. & Unanim. & Original & All types & No plural & No template & Neither\\\midrule
COCO & GDINO-T & 23.42 & 22.63 & 22.44 & 21.77 & 9.47 & 10.14 & 7.79 & 9.67 \\
 & OWL-ViT & 19.79 & 19.79 & 7.65 & 14.99 & 7.66 & 3.40 & 12.05 & 5.69 \\
 & OWLv2 & 5.35 & 5.33 & 3.49 & 4.70 & 1.87 & 1.37 & 2.80 & 2.12 \\
 & YOLO-W-v2 & 6.61 & 6.33 & 5.26 & 7.44 & 2.72 & 2.72 & 3.31 & 3.68 \\
\midrule
LVIS & GDINO-T & 29.91 & 29.52 & 26.56 & 28.33 & 16.80 & 17.90 & 11.97 & 12.35 \\
 & OWL-ViT & 12.23 & 12.21 & 5.08 & 8.76 & 6.68 & 3.48 & 7.51 & 3.94 \\
 & OWLv2 & 5.13 & 5.13 & 1.93 & 4.06 & 2.73 & 1.28 & 3.38 & 1.77 \\
 & YOLO-W-v2 & 12.98 & 12.98 & 7.80 & 12.67 & 7.05 & 5.18 & 8.01 & 6.34 \\
\bottomrule\end{tabular}
\caption{Class-average best-minus-worst gaps in AP points for the matched evaluation populations. Adjudicated and unanimous filters require both equivalence and naturalness. The four right columns sample exactly two original prompts per class under each type restriction.}\label{tab:controls}
\end{table*}

Across 240 dataset--model--class combinations, mean and worst AP$_{50}$ have
Spearman $\rho=0.95$, whereas mean AP and prompt gap have $\rho=-0.25$.
The accompanying correlation matrix reports the finite sample count for
each pair.

Tables~\ref{tab:types-coco} and~\ref{tab:types-lvis} report the COCO and LVIS
prompt-type means. The hypernym and multihop columns score against the child labels only.

\begin{table}[H]
\centering
\scriptsize
\setlength{\tabcolsep}{2.4pt}
\begin{tabular}{lrrrrr}
\toprule
\multicolumn{6}{c}{Equivalent scope} \\
Model & Canon. & Templ. & Syn. & Plur. & Para. \\
\midrule
GDINO-T & 83.25 & 71.66 & 77.26 & 79.06 & 72.62 \\
OWL-ViT & 62.30 & 66.19 & 57.40 & 46.92 & 66.58 \\
OWLv2 & 79.40 & 79.46 & 71.03 & 76.34 & 79.45 \\
YOLO-W-v2 & 77.84 & 75.45 & 70.21 & 76.25 & 74.17 \\
\midrule
\multicolumn{6}{c}{Conditional and broad scopes} \\
Model & Attr. & Context & Comp. & Hyp. & Multi. \\
\midrule
GDINO-T & 77.99 & 72.38 & 69.85 & 56.03 & 50.70 \\
OWL-ViT & 53.04 & 54.04 & 66.20 & 29.31 & 26.53 \\
OWLv2 & 79.28 & 78.90 & 79.53 & 29.93 & 27.55 \\
YOLO-W-v2 & 74.77 & 74.86 & 72.17 & 52.26 & 47.88 \\
\bottomrule
\end{tabular}
\caption{\coco\ mean \apfifty$\times100$ by prompt type. Canon.=canonical;
Templ.=template; Syn.=synonym; Plur.=plural; Para.=paraphrase;
Attr.=attribute; Comp.=compositional; Hyp.=hypernym; Multi.=multihop.}
\label{tab:types-coco}
\end{table}

\begin{table}[H]
\centering
\scriptsize
\setlength{\tabcolsep}{3.5pt}
\begin{tabular}{lrrrrrr}
\toprule
Model & Canon. & Templ. & Plur. & Para. & Comp. & Multi. \\
\midrule
GDINO-T & 62.33 & 35.49 & 58.59 & 49.98 & 47.49 & 35.30 \\
OWL-ViT & 39.49 & 43.88 & 33.01 & 42.94 & 43.05 & 22.05 \\
OWLv2 & 72.31 & 72.29 & 70.09 & 71.67 & 71.21 & 31.04 \\
YOLO-W-v2 & 53.41 & 48.94 & 48.81 & 47.50 & 39.98 & 24.03 \\
\bottomrule
\end{tabular}
\caption{\lvis\ mean \apfifty$\times100$ by prompt type. Abbreviations follow
Table~\ref{tab:types-coco}.}
\label{tab:types-lvis}
\end{table}

\section{LOO and Prompt-Dropout Diagnostics}
Table~\ref{tab:loo} quantifies dependence on a single prompt.
Table~\ref{tab:dropout} reports the 50\% prompt-retention setting; 75\% and
100\% results and all 20-repeat raw samples are retained in the evidence
package. Prompt dropout resamples saved prompt outputs and does not introduce
stochastic detector inference.

\begin{table}[H]
\centering
\scriptsize
\setlength{\tabcolsep}{4.4pt}
\begin{tabular}{llrr}
\toprule
Dataset & Model & $I^W$ & $I^G$ \\
\midrule
\coco & GDINO-T & 11.28 & 11.87 \\
 & OWL-ViT & 12.55 & 12.67 \\
 & OWLv2 & 3.62 & 3.67 \\
 & YOLO-W-v2 & 1.83 & 2.32 \\
\midrule
\lvis & GDINO-T & 12.77 & 14.96 \\
 & OWL-ViT & 7.12 & 7.68 \\
 & OWLv2 & 2.89 & 3.46 \\
 & YOLO-W-v2 & 5.98 & 8.36 \\
\bottomrule
\end{tabular}
\caption{LOO influence in \apfifty\ points. $I^W$ is the largest increase in
worst-prompt AP after removing one prompt; $I^G$ is the largest gap reduction.}
\label{tab:loo}
\end{table}

\begin{table}[H]
\centering
\scriptsize
\setlength{\tabcolsep}{1.0pt}
\begin{tabular}{llrrr}
\toprule
Dataset & Model & M-WBF & Cons. & LSF \\
\midrule
\coco & GDINO-T & $76.25\pm0.31$ & $78.84\pm0.33$ & $79.29\pm0.29$ \\
 & OWL-ViT & $66.07\pm0.24$ & $66.49\pm0.22$ & $66.46\pm0.27$ \\
 & OWLv2 & $78.97\pm0.13$ & $79.01\pm0.12$ & $78.91\pm0.10$ \\
 & YOLO-W-v2 & $75.19\pm0.20$ & $75.45\pm0.21$ & $75.41\pm0.22$ \\
\midrule
\lvis & GDINO-T & $52.39\pm0.90$ & $54.38\pm0.88$ & $54.52\pm0.83$ \\
 & OWL-ViT & $41.25\pm0.48$ & $42.18\pm0.52$ & $42.16\pm0.48$ \\
 & OWLv2 & $68.93\pm0.33$ & $68.75\pm0.31$ & $68.88\pm0.32$ \\
 & YOLO-W-v2 & $47.35\pm0.61$ & $47.78\pm0.66$ & $47.75\pm0.62$ \\
\bottomrule
\end{tabular}
\caption{Prompt dropout at 50\% retention: mean \apfifty$\times100\pm$
standard deviation over 20 repeats. Cons.=Prompt Consensus.}
\label{tab:dropout}
\end{table}

\section{Threshold and All-Class Diagnostics}
The score-threshold sweep evaluates $0.001$, $0.005$, $0.01$, $0.025$, $0.05$,
$0.1$, $0.2$, $0.3$, and $0.5$ for every model and dataset. Because AP ranks
detections, threshold mainly changes the retained candidate pool and
threshold-dependent precision, recall, and false positives. The low-threshold
optimum is stable for Grounding DINO and OWLv2, whereas YOLO-World-v2 loses
recall sooner. We therefore use a common threshold of 0.001 rather than tuning
per model on the test set. Released sweep tables contain 36 rows per dataset.

The canonical all-class COCO protocol has one prompt per class and is not a
lexical-distribution metric. Grounding DINO-Tiny obtains 58.99 AP$_{50}$ and
OWL-ViT obtains 50.05 AP$_{50}$. These values are lower than the positive
protocol means because every image--class pair is queried, including absent
categories. The protocol is retained as a false-positive stress test and should
not be compared numerically with prompt gaps.

\paragraph{Absent-category wording pilot.}
On 100 held-out COCO images, we query OWL-ViT only for classes absent from
the image annotations. A query counts as positive if it returns at least one
box at score $\geq 0.10$. The 193 original prompts yield 1,186 above-threshold boxes in 649 of
17,947 queries (66.08 boxes per 1,000 queries); the 187 adjudicated prompts
yield 1,167 in 638 of 17,362 (67.22 per 1,000). Among 2,799 eligible
image--class pairs, 176 have different binary outcomes across the original
wordings and 170 across the adjudicated wordings. Class-average rates are
6.23\% and 6.03\%, respectively. The query rate weights prompts; the flip rate
first weights eligible images within a class and then gives each class equal
weight. Neither is an AP score. The smoke images belong to the pilot and
are counted only once.

\section{Computational and Environment Details}
The four-detector positive runs use NVIDIA L40 GPUs with 46,068 MiB reported memory,
automatic mixed precision, and TensorFloat-32. The SHiNe/Detic
environment~\cite{liu2024shine,zhou2022detic} uses Python 3.10, PyTorch
2.0.1+cu118, CUDA 11.8, cuDNN 8.7, Detectron2 v0.6 at commit
\texttt{d1e0456}, and the official SHiNe commit \texttt{825a6f7}. The
Detectron2 CUDA extension was compiled against the environment's CUDA 11.8
toolchain and validated on the L40. Environment freezes and GPU identities are
retained with the evidence package.

Table~\ref{tab:compute} gives estimates extrapolated from measured pilot
timings, not cloud bills. COCO includes four positive-protocol and two
canonical all-class cells; LVIS includes four positive-protocol cells.
Aggregation uses saved predictions and requires no GPU.

\begin{table}[H]
\centering
\scriptsize
\setlength{\tabcolsep}{2.3pt}
\begin{tabular}{lrrrr}
\toprule
Stage & Cells & GPU h & Wall h & Cost \\
\midrule
COCO inference & 6 & 48.96 & 12.24 & \$38.68 \\
LVIS extension & 4 & 9.55 & 2.39 & \$7.55 \\
Combined & 10 & 58.52 & 14.63 & \$46.23 \\
\bottomrule
\end{tabular}
\caption{Pilot-calibrated detector resource estimates at \$0.79 per L40-hour.
Wall time assumes four ideally balanced workers and excludes setup, transfer,
synchronization, and metric collection.}
\label{tab:compute}
\end{table}

Multi-prompt methods require $K$ detector passes for $K$ prompts, which is
their dominant deployment cost. On the 1,000-image Detic head-to-head, CPU
aggregation takes 102.9 s for max-score, 112.5 s for Mean-WBF, and 105.4 s for
the primary Prompt Consensus setting. The four-prompt diagnostic therefore has
approximately four-pass detector cost even though its fusion step is modest.
The official SHiNe classifier is a one-pass method.
The separate OWL-ViT negative pilot ran on Apple MPS using the base-patch32
checkpoint; its negative-query loop took 1,600 s. It is not included in the
L40 estimates.

The LSF weights in main-paper Sec.~5 use unlabeled box agreement; only
selection among the 27 settings uses labeled COCO calibration AP.

\textbf{Which classes enter SHiNe?}
All official and aggregation comparisons in the main paper use 30 classes.
In the four-prompt experiment the eight active classes received hierarchy
enrichment in constructing the fixed nexus; the remaining 22 are controls.

Table~\ref{tab:shine-sub} reports paired gap changes within each group. The
bootstrap preserves the LVIS frequency strata; the active group consists of
eight rare classes, while the controls comprise 11 common and 11 frequent
classes. Intervals are pointwise; a control interval including zero does not
establish equality. The fixed-nexus experiment varies the leaf string while
holding the hierarchy construction fixed.
\begin{center}\small
\begin{tabular}{lrrl}\toprule
Population & Classes & $\Delta$ gap & 95\% CI (points)\\\midrule
All & 30 & $-1.64$ & $[-3.75,-0.20]$\\
Active & 8 & $-6.25$ & $[-14.18,-0.96]$\\
Inactive & 22 & $+0.04$ & $[-0.001,0.113]$\\\bottomrule
\end{tabular}
\captionof{table}{Change in the four-prompt gap (SHiNe fixed nexus minus Detic) in
AP$_{50}$ points. Negative values denote a smaller gap.}\label{tab:shine-sub}
\end{center}

\section{Class support and visual examples}
Tables~\ref{tab:class-COCO} and~\ref{tab:class-LVIS} list the five largest gaps for each detector
on each dataset, with the corresponding positive-image and target-instance
counts. Intervals are the 2.5th and 97.5th percentiles of 1,000 within-class
image-bootstrap gap estimates. Resampling an image preserves its objects and
all evaluated wordings. These intervals do not adjust for selecting the
largest estimates or for multiple comparisons. With five images, an interval
can be wide even when the observed gap is large. The accompanying CSV includes
all tied best and worst strings, as well as the 240 estimates and intervals.

\begin{figure}[H]\centering
\begin{minipage}[t]{.322\linewidth}\centering
\includegraphics[width=\linewidth]{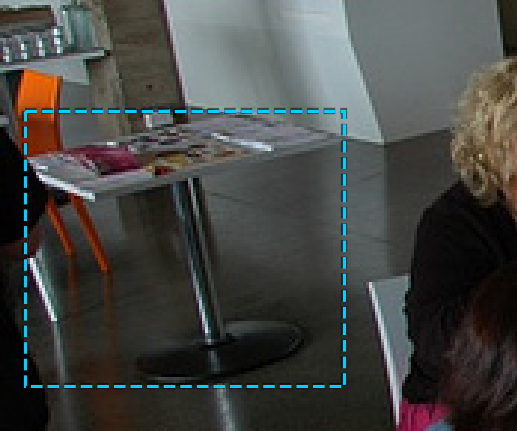}\par
{\footnotesize (a) \emph{dining table}\par 0.071\par}\end{minipage}\hfill
\begin{minipage}[t]{.322\linewidth}\centering
\includegraphics[width=\linewidth]{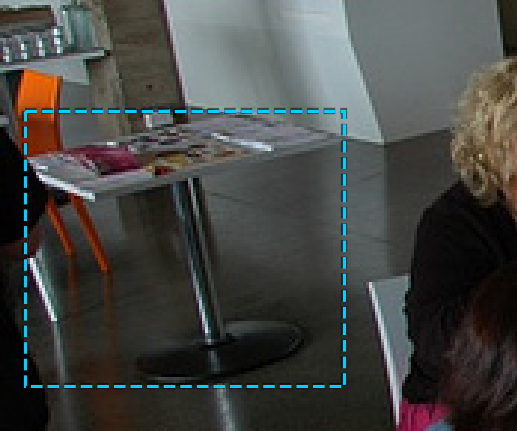}\par
{\footnotesize (b) \emph{dinner table}\par 0.095\par}\end{minipage}\hfill
\begin{minipage}[t]{.322\linewidth}\centering
\includegraphics[width=\linewidth]{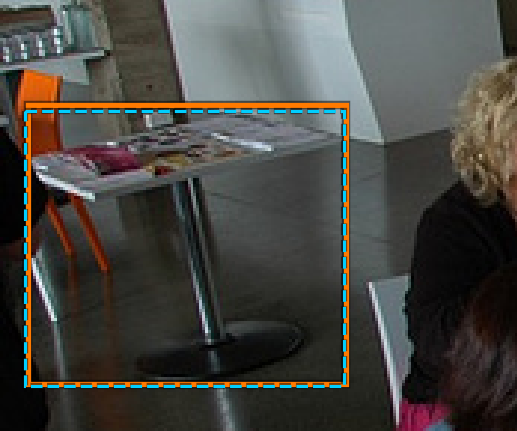}\par
{\footnotesize (c) \emph{dining tables}\par 0.173\par}\end{minipage}
\caption{\textbf{Same proposal, different detection outcome.} OWLv2, COCO image 306139.
Dashed cyan marks the annotated table. The values are raw scores for the same box.
Only \emph{dining tables} exceeds the 0.10 threshold; its matched prediction appears in solid orange.}
\label{fig:dining}\end{figure}

Figure~\ref{fig:dining} follows one annotated dining table under three equivalent prompts.
Changing the wording changes the score of a single proposal without changing its coordinates.

\noindent\begin{minipage}[t]{\textwidth}\centering\footnotesize
\setlength{\tabcolsep}{2.6pt}
\begin{tabular}{@{}llrrrr@{}}\toprule
Model & Class & Img. & Inst. & Gap & 95\% CI\\\midrule
GDINO-T & traffic light & 35 & 109 & 48.8 & [42.8, 55.0] \\
 & car & 93 & 343 & 47.4 & [42.7, 50.6] \\
 & sheep & 10 & 44 & 44.4 & [27.7, 54.3] \\
 & person & 481 & 1952 & 38.4 & [35.5, 41.0] \\
 & cow & 11 & 49 & 37.0 & [26.8, 44.7] \\
\midrule
OWL-ViT & tv & 40 & 56 & 47.7 & [32.3, 57.2] \\
 & dog & 37 & 43 & 40.0 & [27.9, 49.0] \\
 & toilet & 30 & 39 & 31.7 & [20.8, 47.1] \\
 & zebra & 17 & 69 & 31.3 & [21.5, 36.4] \\
 & airplane & 20 & 27 & 30.6 & [15.1, 39.5] \\
\midrule
OWLv2 & laptop & 36 & 50 & 24.2 & [14.3, 31.0] \\
 & traffic light & 35 & 109 & 20.0 & [9.9, 30.1] \\
 & dining table & 94 & 116 & 10.0 & [4.5, 14.8] \\
 & boat & 19 & 60 & 9.8 & [4.2, 15.8] \\
 & car & 93 & 343 & 9.8 & [5.6, 14.5] \\
\midrule
YOLO-W-v2 & cell phone & 34 & 44 & 20.9 & [9.9, 30.1] \\
 & car & 93 & 343 & 15.4 & [11.5, 17.8] \\
 & person & 481 & 1952 & 12.3 & [9.7, 15.7] \\
 & couch & 39 & 53 & 11.8 & [3.9, 16.8] \\
 & book & 38 & 164 & 10.2 & [4.8, 18.4] \\
\bottomrule\end{tabular}\captionof{table}{COCO: five largest original AP$_{50}$ prompt gaps per model, in points. Img.\ and Inst.\ give the number of positive images and instances; intervals are pointwise.}\label{tab:class-COCO}\end{minipage}\par\medskip
\begin{minipage}[t]{\textwidth}\centering\footnotesize
\setlength{\tabcolsep}{2.6pt}
\begin{tabular}{@{}llrrrr@{}}\toprule
Model & Class & Img. & Inst. & Gap & 95\% CI\\\midrule
GDINO-T & radio receiver & 9 & 10 & 65.0 & [45.1, 77.8] \\
 & stepladder & 5 & 5 & 57.5 & [27.9, 86.9] \\
 & traffic light & 114 & 469 & 52.0 & [48.5, 55.6] \\
 & pillow & 95 & 293 & 51.6 & [47.2, 54.2] \\
 & kitchen table & 5 & 5 & 50.1 & [13.2, 53.4] \\
\midrule
OWL-ViT & kettle & 5 & 9 & 47.1 & [39.9, 76.8] \\
 & stepladder & 5 & 5 & 29.2 & [4.5, 65.2] \\
 & horse & 107 & 263 & 23.9 & [17.7, 30.0] \\
 & radio receiver & 9 & 10 & 21.8 & [6.7, 54.3] \\
 & laptop computer & 97 & 124 & 21.7 & [15.0, 28.2] \\
\midrule
OWLv2 & radio receiver & 9 & 10 & 20.1 & [4.4, 44.1] \\
 & binder & 6 & 15 & 17.1 & [9.4, 41.9] \\
 & legume & 5 & 242 & 13.6 & [6.1, 25.7] \\
 & papaya & 5 & 90 & 13.5 & [4.4, 20.0] \\
 & kitchen table & 5 & 5 & 9.6 & [0.0, 17.0] \\
\midrule
YOLO-W-v2 & stepladder & 5 & 5 & 73.3 & [36.4, 100.0] \\
 & binder & 6 & 15 & 37.5 & [5.5, 66.8] \\
 & shampoo & 6 & 21 & 29.2 & [21.9, 39.3] \\
 & egg roll & 5 & 28 & 25.9 & [16.5, 46.9] \\
 & ashtray & 11 & 18 & 20.1 & [8.1, 37.9] \\
\bottomrule\end{tabular}\captionof{table}{LVIS: five largest original AP$_{50}$ prompt gaps per model, in points. Img.\ and Inst.\ give the number of positive images and instances; intervals are pointwise.}\label{tab:class-LVIS}\end{minipage}\par\smallskip\nointerlineskip

\begin{figure}[H]\centering
\begin{minipage}[t]{.322\linewidth}\centering
\includegraphics[width=\linewidth]{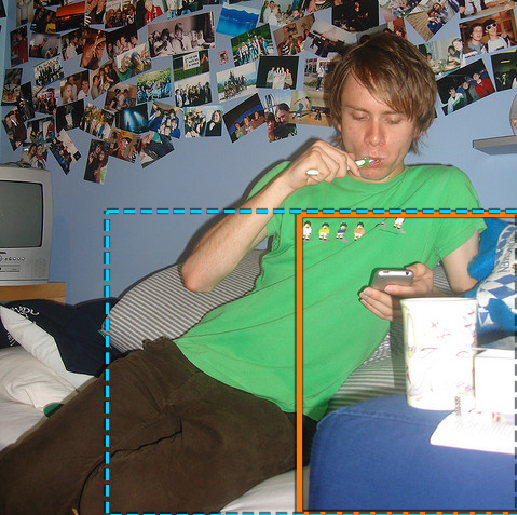}\par
{\footnotesize (a) \emph{a bed}\par Score: 0.164\par}\end{minipage}\hfill
\begin{minipage}[t]{.322\linewidth}\centering
\includegraphics[width=\linewidth]{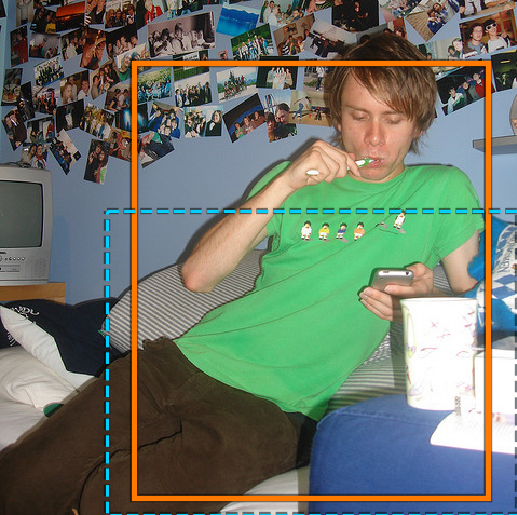}\par
{\footnotesize (b) \emph{a visible bed}\par Score: 0.125\par}\end{minipage}\hfill
\begin{minipage}[t]{.322\linewidth}\centering
\includegraphics[width=\linewidth]{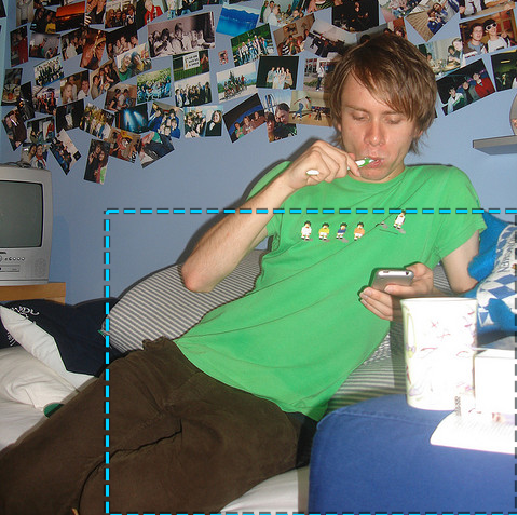}\par
{\footnotesize (c) \emph{beds}\par No match\par}\end{minipage}
\caption{\textbf{Different box extent and a missed match.} OWL-ViT, COCO image 218439.
The first two prompts match the same annotated bed, but their orange boxes overlap each other at only 0.322 IoU. For \emph{beds}, no prediction passes both the score and matching thresholds for this instance.}
\label{fig:bed}\end{figure}

\begin{figure}[H]\centering
\begin{minipage}[t]{.322\linewidth}\centering
\includegraphics[width=\linewidth]{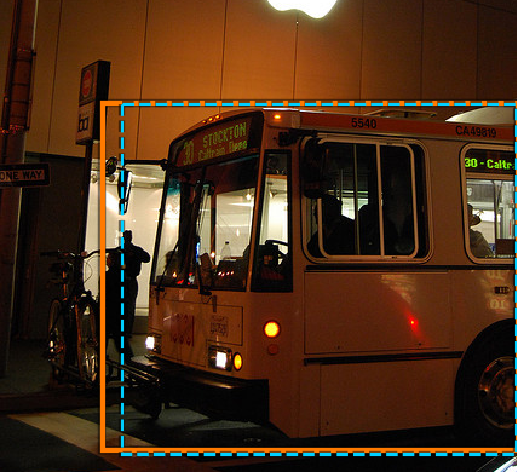}\par
{\footnotesize (a) \emph{bus}\par Score: 0.932\par}\end{minipage}\hfill
\begin{minipage}[t]{.322\linewidth}\centering
\includegraphics[width=\linewidth]{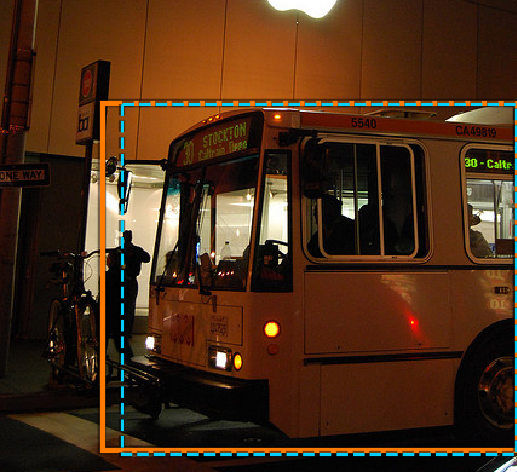}\par
{\footnotesize (b) \emph{the bus in the image}\par Score: 0.946\par}\end{minipage}\hfill
\begin{minipage}[t]{.322\linewidth}\centering
\includegraphics[width=\linewidth]{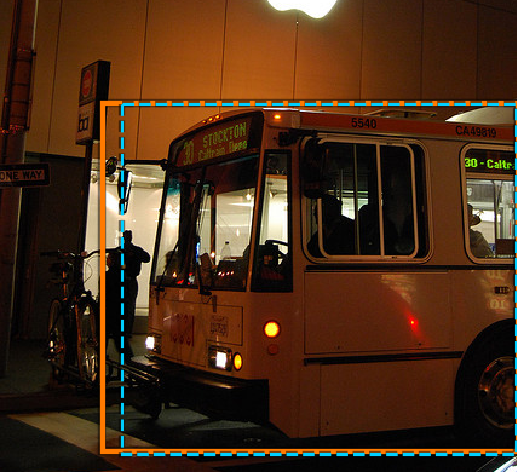}\par
{\footnotesize (c) \emph{a photo of a bus}\par Score: 0.977\par}\end{minipage}
\caption{\textbf{Stable localization despite wording changes.} YOLO-World-v2, COCO image 306136. Across all five equivalent prompts evaluated for this bus, the minimum pairwise IoU between matched predictions is 0.995. Three of the five wordings are shown.}
\label{fig:bus}\end{figure}

\noindent\textbf{Box extent and detection stability.}
On the partly occluded bed (Figure~\ref{fig:bed}), wording changes both the extent of the matched box and whether a match exists. The bus (Figure~\ref{fig:bus}) provides a stable comparison.
Within each figure the image crop, target and detector are fixed. Dashed cyan marks the target; solid orange marks its matched prediction (score $\geq 0.10$, IoU with the target $\geq 0.50$). Other detections are omitted.
All nine displayed prompts were retained by the human audit. These examples illustrate possible outcomes, not their frequency.

\section{Using the accompanying files}
The ZIP (\href{https://www.dropbox.com/scl/fi/kjtkwrjmv06bvfg9iettj/02112_supp.zip?rlkey=obxvgdb4o023rs0i2xymu98qg&st=e2zc1e1x&dl=0}{link here}) includes 620 candidate records, the rater instructions and decisions,
and the numerical results underlying the tables. Its README identifies
the files, units, release terms and contact for corrections. Report the
dataset and stable prompt ID when proposing a change.

\end{document}